\documentclass[preprint,authoryear,12pt]{elsarticle}

\usepackage[utf8]{inputenc}
\usepackage{longtable}
\usepackage[T1]{fontenc}
\usepackage{amsmath,amssymb,amsthm}
\usepackage{graphicx}
\usepackage{booktabs}
\usepackage{multirow}
\usepackage{array}
\usepackage{xcolor}
\usepackage{hyperref}
\usepackage{pgfplots}
\usepackage{algorithm}
\usepackage{algorithmic}
\usepackage{tabularx}
\usepackage{makecell}
\usepackage{tikz}
\usepackage{placeins}   
\usepackage{enumerate}
\usepackage{pgfplots}
\usepackage{pgfplotstable}
\pgfplotsset{compat=1.18}
\usepgfplotslibrary{groupplots}
\usepackage[left=2.2cm, right=2.2cm, top=2.5cm, bottom=2.5cm]{geometry}
\usepackage{placeins}   
\usepackage{enumitem}
\usetikzlibrary{shapes.geometric, arrows.meta, positioning,
                fit, backgrounds, calc}

\providecommand{\doi}[1]{%
    \href{https://doi.org/#1}{doi:\,\nolinkurl{#1}}%
}
\providecommand{\Comment}[1]{\COMMENT{#1}}

\let\cite\citep

\hypersetup{colorlinks=true,linkcolor=blue,citecolor=blue,urlcolor=blue}

\tikzset{
  block/.style={rectangle, draw=black!70, fill=blue!10,
                text width=2.6cm, text centered,
                rounded corners=3pt, minimum height=0.75cm,
                font=\scriptsize},
  greenblock/.style={rectangle, draw=green!60!black, fill=green!12,
                text width=2.6cm, text centered,
                rounded corners=3pt, minimum height=0.75cm,
                font=\scriptsize\bfseries},
  redblock/.style={rectangle, draw=red!60!black, fill=red!10,
                text width=2.6cm, text centered,
                rounded corners=3pt, minimum height=0.75cm,
                font=\scriptsize},
  opblock/.style={rectangle, draw=orange!70!black, fill=orange!12,
                text width=2.3cm, text centered,
                rounded corners=3pt, minimum height=0.65cm,
                font=\scriptsize},
  arrow/.style={-Stealth, thick, draw=black!70},
  garrow/.style={-Stealth, thick, draw=green!60!black},
  rarrow/.style={-Stealth, thick, draw=red!60!black},
  dashed arrow/.style={-Stealth, dashed, draw=gray!70},
}

\begin{document}
\begin{frontmatter}
\title{Green ACO: An Energy-Aware Ant Colony Optimization \\ Framework for MAX-SAT}

\author[esi]{Maroua Ogab, Nazim Abdeerhmane Aouni,
             Rayane Rahmat Errahmane Smara, Nour El Imane Elbar,
             Zakaria Soualah Mohammed, Malika Bessedik}

\address[esi]{Higher National School of Computer Science (ESI), Algiers, Algeria\\[2pt]
\normalfont\footnotesize \{mm\_ogab, mn\_aouni, mr\_smara, mn\_elbar,
mz\_soualahmohammed, m\_bessedik\}@esi.dz}

\begin{abstract}
NP-hard combinatorial optimization problems demand heavy computation, and this raises real environmental concerns, especially in regions with carbon-intensive power grids. This paper presents Green ACO, an energy-aware Ant Colony Optimization framework for the MAXIMUM SATISFIABILITY (MAX-SAT) problem. At its core is an Expected Improvement per Joule (EI/J) scheduler. At every iteration, it picks the operator that best balances expected quality gain against expected energy cost. Three operators feed the scheduler: WalkSAT, Focused Variable Neighbourhood Search, and Clause-Restart Greedy, ranging from cheap and surgical to aggressive and costly. An offline dual-regime profiling phase calibrates these operators before the search even begins, and, combined with Thompson sampling, this reduces uninformed early operator choices; online call-count warm-up still occurs. A pre-dispatch guard rejects calls when their estimated energy cost, with a safety multiplier, exceeds the remaining budget. Sparse pheromone updates, pheromone caching across budgets, and stagnation-based early stopping trim energy use further, without hurting solution quality. The framework also incorporates a budget predictor that estimates the energy budget required for unseen instances. On four hard MAX-SAT reference instances, Green ACO reaches an average quality of 98.44\%. This beats the standard ACO baselines by close to 3 points, while using only 986~J against 4,171 to 6,610~J for the strongest competitors, a six-and-a-half-fold gain in quality per joule. Under the Algerian grid's carbon intensity, this cuts CO$_2$e emissions by approximately 85\% compared with the most energy-hungry baseline.
\end{abstract}


\end{frontmatter}
\section{Introduction}
\label{sec:intro}

\subsection{Problem Definition}

MAX-SAT is the optimisation version of Boolean Satisfiability. Given a propositional formula in conjunctive normal form, the goal is to find the truth assignment that satisfies as many clauses as possible. MAX-SAT inherits the NP-hardness of SAT, since SAT is a special case of it, and no known algorithm solves it in polynomial time in the general case. Exact methods do not scale either: even mid-sized industrial instances become intractable quickly. Metaheuristic search remains the practical choice once a formula reaches tens of thousands of variables and clauses, but that choice comes with a cost of its own. Reaching high-quality solutions on large instances often means running longer, evaluating more candidates, and consuming more energy along the way. This tension between solution quality and energy consumption is exactly what green computing tries to address, and it sits at the center of the problem tackled in this paper. Instances of this scale show up often in formal verification, automated planning, knowledge compilation, and the encoding of decision-tree classifiers into propositional logic~\cite{hoos2004stochastic,li2009maxsat}.

\subsection{Standard Metaheuristics}

Several metaheuristics have been applied to MAX-SAT with success. Genetic Algorithms, Ant Colony Optimisation, Simulated Annealing, and Particle Swarm Optimisation all balance exploration and exploitation reasonably well over large search spaces~\cite{dorigo1997aco,balint2012random,smyth2003,selman1994noise}. What most of these methods ignore is the energy their search actually costs. Abdelhafez et al.~\cite{abdelhafez2019} found that algorithmic and parameter choices alone can shift energy use by a factor of two to four, even when solution quality barely changes. A large part of the energy spent by these solvers buys almost nothing in return.

\subsection{Green Computing}

This gap is what motivates treating energy as a real objective, not an afterthought. Green computing has grown into an active field on its own, pushed mainly by the environmental cost of large-scale computing. Designing a green metaheuristic means rethinking the search itself, rather than just limiting its runtime from outside. Early stopping criteria can cut a run short once further search stops paying off. Reducing the search space can avoid wasted evaluations altogether. Eco-efficient operators can be chosen or built to deliver quality gains at a lower energy cost per step. These are the general levers this paper draws on, and the goal is to combine them into a solution where energy awareness is built into the algorithm's own decisions, not added on afterward.

\subsection{Green Metaheuristics}

Work that explicitly combines green computing with metaheuristic search is still rare. Verdecchia et al.~\cite{verdecchia2023} point to hardware-level techniques, approximate computing, and algorithm-level resource management as the three main levers for cutting energy use. Jamil and Kor~\citeyearpar{jamil2022energy} show, across several nature-inspired algorithms, that population size and operator choice drive most of the energy consumption. Closer to this paper, Alba, Harada and Luque~\citeyearpar{alba2026} proposed the Expected Improvement per Joule (EI/J) criterion. It selects operators based on the ratio of expected quality gain to expected energy cost, and was tested on Genetic Algorithms, Particle Swarm Optimisation, and Iterated Local Search on generic combinatorial problems. Neither Ant Colony Optimisation nor MAX-SAT had been studied under this criterion before, and resource-constrained SAT solvers in the literature still reason mostly in terms of runtime limits, not energy~\cite{smyth2003}. This gap is what the present paper addresses.

\subsection{Contribution and Originality}

This paper builds upon the Expected Improvement per Joule (EI/J) criterion proposed by~\cite{alba2026}, originally developed for Genetic Algorithms (GA), Particle Swarm Optimisation (PSO), and Iterated Local Search (ILS), and extends it to Ant Colony Optimisation (ACO) for the MAX-SAT problem. Unlike these algorithms, ACO relies on a pheromone structure that accumulates search experience across iterations, while its search operators exhibit substantially different computational costs and optimisation behaviours. These characteristics require a dedicated energy-aware scheduling strategy, building on adaptive operator-selection work~\cite{DaCosta2008}, rather than a direct application of the original EI/J framework.

To address these challenges, this paper makes the following contributions:

\begin{itemize}
    \item \textbf{Energy-aware ACO for MAX-SAT.} We propose \emph{Green ACO}, the first Ant Colony Optimisation framework for MAX-SAT that integrates Expected Improvement per Joule (EI/J) scheduling to maximize solution quality under constrained energy budgets.

    \item \textbf{Cold-start-aware operator scheduling.} We introduce a dual-regime offline profiling strategy that characterizes operator behaviour under both easy and difficult search regimes. Combined with Thompson sampling, this initialization provides informative priors that reduce uninformed early scheduling decisions; online call-count warm-up remains.

    \item \textbf{Instance-aware energy budget prediction.} We propose a lightweight budget predictor that estimates the energy budget required for previously unseen MAX-SAT instances from inexpensive structural features. Since the predictor is independent of Green ACO, it can be integrated into a broad class of energy-aware metaheuristics to support budget initialization and resource allocation.

    \item \textbf{Budget-aware execution.} We introduce a pre-dispatch anti-overrun guard that rejects calls whose estimated energy cost, with a safety multiplier, exceeds the remaining budget.

    \item \textbf{Lightweight energy-saving mechanisms.} We further reduce computational energy consumption through sparse pheromone updates, pheromone caching across budget levels, and stagnation-based early stopping while preserving solution quality.
\end{itemize}

Green ACO is evaluated on 50 hard MAX-SAT instances, most from the SAT 2024 benchmark suite. Comparative results against existing ACO variants are reported on four hard reference instances; ablation and operator-validation analyses assess the proposed mechanisms.

\subsection{Paper Organisation}

The remainder of this paper is organized as follows. Section~\ref{sec:problem} introduces the MAX-SAT problem and the energy model adopted throughout this work. Section~\ref{sec:method} presents the proposed Green ACO framework in full. Section~\ref{sec:experiments} describes the experimental methodology, benchmark instances, parameter tuning procedure, and presents the comparative, statistical, and ablation results together with their discussion. Finally, Section~\ref{sec:conclusion} concludes the paper and outlines directions for future research.
\section{Problem Formulation and Energy-Aware Optimization Model}
\label{sec:problem}

\subsection{MAX-SAT Formal Definition}
\label{subsec:maxsat_def}

Let $V=\{x_1,x_2,\ldots,x_n\}$ be a finite set of Boolean variables.
A \emph{literal} is either a variable $x_i$ or its negation
$\bar{x}_i$. A \emph{clause} $C_j$ is a disjunction of literals.
A CNF formula is defined as

\begin{equation}
\varphi=C_1\land C_2\land\cdots\land C_m.
\label{eq:formula}
\end{equation}

An \emph{assignment} is a function

\[
\mathbf{x}:V\rightarrow\{0,1\},
\]
\noindent
and, for conciseness, is denoted by
$\mathbf{x}\in\{0,1\}^n$.

Given a CNF formula $\varphi$, the \textsc{Max-Sat} problem consists of
finding the assignment that satisfies the largest possible number of
clauses:

\begin{equation}
\max_{\mathbf{x}\in\{0,1\}^n}
\sum_{j=1}^{m}
\mathbf{1}[C_j(\mathbf{x})=\top].
\label{eq:maxsat_obj}
\end{equation}

The \emph{fitness} of an assignment corresponds to the number of
satisfied clauses:

\begin{equation}
f(\mathbf{x})
=
\sum_{j=1}^{m}
\mathbf{1}[C_j(\mathbf{x})=\top],
\qquad
f\in\{0,\ldots,m\}.
\label{eq:fitness}
\end{equation}

To facilitate comparisons across benchmark instances containing
different numbers of clauses, solution quality is normalized as

\begin{equation}
q(\mathbf{x})
=
\frac{f(\mathbf{x})}{m}\times100\%.
\label{eq:quality}
\end{equation}

The set of unsatisfied clauses under assignment $\mathbf{x}$ is

\[
\mathcal{U}(\mathbf{x})
=
\{j:C_j(\mathbf{x})=\bot\}.
\]

Given a variable $x_v$, flipping its Boolean value generates the
neighboring assignment $\mathbf{x}^{(v)}$. The corresponding variation
in fitness is

\[
\Delta f_v
=
f(\mathbf{x}^{(v)})
-
f(\mathbf{x}).
\]

This formulation defines the optimization objective independently of
the optimization algorithm. Green ACO searches this solution space by
constructing candidate assignments using an Ant Colony Optimization
framework, presented next.

\subsection{ACO Pheromone Model for MAX-SAT}
\label{subsec:pheromone}

Given the optimization problem defined above, candidate assignments are
constructed using an Ant Colony Optimization (ACO) framework. Each ant
incrementally assigns Boolean values to variables according to a
pheromone model that combines previously acquired search experience with
problem-specific heuristic information.

The ACO pheromone matrix is

\[
\tau
\in
[\tau_{\min},\tau_{\max}]^{n\times2},
\]
\noindent
where $\tau[v][b]$ represents the learned desirability of assigning
Boolean value $b\in\{0,1\}$ to variable $v$.

Assignments are generated according to the following construction rule:

\begin{equation}
\mathbf{x}_v=
\begin{cases}
\arg\max_b
\tau[v][b]\eta[v][b]^{\beta},
&
\text{with probability }q_0,
\\
\text{sample proportional to }
\tau[v][b]\eta[v][b]^{\beta},
&
\text{otherwise.}
\end{cases}
\label{eq:construction}
\end{equation}
where

\[
\eta[v][b]
=
|\{j:\text{literal}(v,b)\in C_j\}|
\]
\noindent
is the heuristic information measuring how many clauses are immediately
supported by assigning value $b$ to variable $v$.

The parameter $\beta\ge0$ controls the influence of heuristic
information, whereas $q_0\in[0,1]$ balances exploitation and
exploration during solution construction.

All pheromone values are initialized uniformly to
$\tau_0\in(0,1)$.
As the search progresses, pheromone values are reinforced according to
the quality of previously generated solutions, progressively guiding the
search toward more promising regions of the solution space.

While the previous subsections define the optimization objective and the
solution construction mechanism, Green ACO additionally considers the
computational resources required to perform the search. The following
subsection therefore introduces the energy model adopted throughout this
work.

\subsection{Energy Model}
\label{subsec:energy}

Traditional metaheuristics are generally evaluated according to
solution quality and execution time. In contrast, Green ACO explicitly
models computational energy as an optimization resource. This section
defines the energy and carbon models used to quantify the environmental
cost of the search process.

\subsubsection{Energy Consumption}

The energy consumed during one execution of the algorithm is denoted by

\[
E\in\mathbb{R}_{>0},
\]
\noindent
and is expressed in joules (J).

In practice, energy consumption can be measured using hardware performance counters (e.g., Intel RAPL), estimated from execution time and average processor power, or monitored using software energy tracking tools such as CodeCarbon~\cite{courty2023codecarbon}, which is the library used throughout this work.

The total energy consumed during one execution is

\begin{equation}
E_{\text{total}}
=
\sum_{t=1}^{T}
E_t,
\label{eq:energy_total}
\end{equation}
\noindent
where $T$ denotes the total number of executed iterations and $E_t$ represents the energy consumed during iteration $t$.

\subsubsection{Carbon Footprint}

The CO$_2$-equivalent emissions from a computation consuming $E$ joules are

\begin{equation}
\text{CO}_2\text{e}\,[\mu\text{g}]
=
\frac{E\,[\text{J}]}
{3.6\times10^6}
\times
\text{CI}
\left[
\frac{\text{gCO}_2\text{e}}{\text{kWh}}
\right]
\times10^6,
\label{eq:co2}
\end{equation}
\noindent
where CI denotes the carbon intensity of the local electrical grid, expressed in gCO$_2$e/kWh.

Carbon intensity varies significantly across countries according to
their electricity production mix. Consequently, identical computations
may produce substantially different environmental impacts. Illustrative
values are presented in Table~\ref{tab:ci}.

\begin{table}[H]
\centering
\caption{Illustrative Regional Carbon Intensities (gCO$_2$e/kWh). Source:
Low Carbon Power, 2023 data; accessed 17 November 2026.}
\label{tab:ci}
\small
\begin{tabular}{lclc}
\toprule
Region & CI & Region & CI\\
\midrule
Norway (NO) & 31  & France (FR)  & 42  \\
UK (GB)     & 216 & Germany (DE) & 298 \\
USA (US)    & 343 & Algeria (DZ) & 486 \\
China (CN)  & 477 & --           & --  \\
\bottomrule
\end{tabular}
\end{table}
\section{Green ACO: Proposed Method}
\label{sec:method}

\subsection{Global Architecture}
\label{subsec:architecture}

Green ACO extends classical Ant Colony Optimisation for MAX-SAT with a set of energy-aware mechanisms designed to maximize solution quality under constrained energy budgets. The framework consists of an offline initialization phase followed by an iterative optimization loop. During initialization, the offline profiling procedure, described in Section~\ref{subsec:green_components}, calibrates the behaviour of every operator before the search begins. The optimization loop then runs while the available energy budget permits execution. At each iteration, the EI/J scheduler selects an operator, subject to an anti-overrun guard that rejects calls when their estimated energy cost, multiplied by a safety factor, exceeds the remaining budget. After execution, the pheromone matrix is updated using sparse writes and, across successive budget levels, pheromone warm-restart caching. The loop terminates when the energy budget is exhausted, the search has stagnated, or solution quality has reached a plateau, as detailed in Section~\ref{subsec:green_components}.

\begin{figure}[t]
\centering
\resizebox{\columnwidth}{!}{%
\begin{tikzpicture}[node distance=0.65cm and 0.55cm]
  \colorlet{tealfill}  {green!8!white}
  \colorlet{tealstroke}{green!50!black}
  \colorlet{bluefill}  {blue!8!white}
  \colorlet{bluestroke}{blue!60!black}
  \colorlet{amberfill} {orange!12!white}
  \colorlet{amberstroke}{orange!70!black}
  \colorlet{redfill}   {red!8!white}
  \colorlet{redstroke} {red!60!black}
  \colorlet{grayfill}  {black!5!white}
  \colorlet{graystroke}{black!50}
  \colorlet{loopclr}   {black!40}
  \tikzset{
    gacbase/.style={
      rectangle, rounded corners=3pt,
      text centered, font=\scriptsize,
      inner sep=4pt, outer sep=0pt,
      minimum height=0.72cm,
    },
    tealnode/.style={gacbase, text width=#1, fill=tealfill,  draw=tealstroke},
    tealnode/.default=3.0cm,
    bluenode/.style={gacbase, text width=#1, fill=bluefill,  draw=bluestroke},
    bluenode/.default=3.0cm,
    ambernode/.style={gacbase, text width=#1, fill=amberfill, draw=amberstroke},
    ambernode/.default=2.6cm,
    rednode/.style={gacbase, text width=#1, fill=redfill,   draw=redstroke},
    rednode/.default=3.0cm,
    graynode/.style={gacbase, text width=#1, fill=grayfill,  draw=graystroke},
    graynode/.default=3.0cm,
    zone/.style={
      draw, dashed, rounded corners=5pt,
      inner sep=8pt, outer sep=0pt,
      font=\scriptsize\itshape,
    },
    offzone/.style={zone, draw=tealstroke},
    onzone/.style ={zone, draw=bluestroke},
    edgebase/.style={-{Stealth[length=4pt]}, thick,
                     rounded corners=4pt, shorten >=1pt},
    ga/.style={edgebase, draw=black!60},
    gg/.style={edgebase, draw=tealstroke},
    gr/.style={edgebase, draw=redstroke},
    go/.style={edgebase, draw=amberstroke},
    gd/.style={edgebase, dashed, draw=loopclr},
    elab/.style={font=\tiny, fill=white, inner sep=1pt, text=black!70},
    legswatch/.style={rectangle, draw, minimum width=0.28cm, minimum height=0.28cm, inner sep=0pt},
  }

  \node[tealnode=2.8cm] (profiling) {
    \textbf{Dual-regime profiling}\\
    {\tiny easy ($q_0{=}0.5$) + hard ($q_0{=}0.95$)}
  };
  \node[tealnode=3.2cm, right=1.1cm of profiling] (prior) {
    \textbf{Prior seeding (L1--4)}\\
    {\tiny $\mu_{\Delta f},\sigma_{\Delta f},
           \mu_{\ln E},\sigma_{\ln E}\!\to\!$EWMA}
  };
  \draw[gg] (profiling) -- (prior);

  \node[graynode=3.0cm, below=1.5cm of prior, xshift=-1.6cm] (init) {
    \textbf{Initialisation}
  };

  \node[bluenode=4.6cm, below=0.85cm of init, xshift=0.5cm] (sched) {
    \textbf{EI/J scheduler}\quad $o^\star=\arg\max_o \Phi(o)$\\
    {\tiny $\Phi(o)=\dfrac{\Delta f}{E}\times\pi(B),\ \ \pi=\dfrac{B}{B+\mathbb{E}[E]}$
           $\cdot$ Thompson sampling}
  };
  \node[rednode=3.0cm, right=1.4cm of sched] (guard) {
    \textbf{Anti-overrun}\\
    {\tiny reject if $\mathbb{E}[E(o)] > B\cdot\kappa$}
  };
  \draw[ga] (sched) -- (guard);
  \draw[gr] (guard.north) to[bend right=35]
        node[elab, pos=0.5]{reject} (sched.north);

  \node[ambernode=2.6cm, below=1.3cm of sched, xshift=-3.4cm] (walksat) {
    \textbf{WalkSAT}\\
    {\tiny strategic flip}
  };
  \node[ambernode=2.6cm, below=1.3cm of sched] (vns) {
    \textbf{Focused VNS}\\
    {\tiny $k$-flip, pheromone}
  };
  \node[ambernode=2.6cm, below=1.3cm of sched, xshift=3.4cm] (restart) {
    \textbf{Clause restart}\\
    {\tiny partial reconstruction}
  };
  \draw[go] (sched) -- (walksat);
  \draw[go] (sched) -- (vns);
  \draw[go] (sched) -- (restart);

  \node[tealnode=7.0cm, below=1.3cm of vns] (ewma2) {
    \textbf{EWMA update (Operator/Stats)}\\
    {\tiny $\mu_{\Delta f},\sigma_{\Delta f},\mu_{\ln E},\sigma_{\ln E}
           \leftarrow \alpha\cdot\text{prev}+(1-\alpha)\cdot\text{obs}$}
  };
  \draw[ga] (walksat) -- (ewma2);
  \draw[ga] (vns)     -- (ewma2);
  \draw[ga] (restart) -- (ewma2);

  \node[tealnode=7.0cm, below=0.85cm of ewma2] (pher) {
    \textbf{Sparse pheromone update}\ $O(k)$, $k\ll n$\\
    {\tiny frugal skip $\cdot$ adaptive evaporation $\cdot$ cache;\ \
           min-budgets $\tau\leftarrow(1-\rho)\tau+\delta_{gain}$}
  };
  \draw[gg] (ewma2) -- (pher);

  \node[rednode=5.2cm, below=0.85cm of pher] (stag) {
    \textbf{Stagnation / plateau check}\\
    {\tiny $\Delta f_{W=10}<\varepsilon$ or quality $\ge 99.5\%$?}
  };
  \draw[ga] (pher) -- (stag);

  \node[rednode=5.2cm, below=0.85cm of stag] (stop) {
    \textbf{Stop --- return $x^\star$}\\
    {\tiny quality $\cdot$ energy $\cdot$ CO$_2$e $\cdot$ score}
  };
  \draw[gr] (stag) -- node[elab]{yes} (stop);

  \coordinate (looplane) at ($(stag.west)+(-1.0,0)$);
  \draw[gd] (stag.west) -- (looplane) |-
        node[elab, pos=0.22, rotate=90]{no} (sched.west);

  \draw[ga] (init) -- (sched);

  \draw[gd] (prior.south) -- ++(0,-0.25) -| (init.north)
        node[elab, pos=0.85]{seed};

  \begin{scope}[on background layer]
    \node[offzone,
          fit=(profiling)(prior),
          label={[tealstroke,font=\scriptsize\itshape]
                 north west:Offline phase}]
         (offbox) {};
    \node[onzone,
          fit=(init)(sched)(guard)(walksat)(vns)(restart)
              (ewma2)(pher)(stag)(stop)(looplane),
          label={[bluestroke,font=\scriptsize\itshape]
                 north east:Online loop --- while budget $>0$}]
         (onbox) {};
  \end{scope}

  \node[legswatch, fill=tealfill,  draw=tealstroke,
        below=0.9cm of onbox.south, xshift=-4.6cm] (l1) {};
  \node[right=2pt of l1, font=\tiny] (l1t) {Eco-efficiency};
  \node[legswatch, fill=bluefill,  draw=bluestroke,  right=0.5cm of l1t] (l2) {};
  \node[right=2pt of l2, font=\tiny] (l2t) {EI/J scheduler};
  \node[legswatch, fill=amberfill, draw=amberstroke, right=0.5cm of l2t] (l3) {};
  \node[right=2pt of l3, font=\tiny] (l3t) {Operators};
  \node[legswatch, fill=redfill,   draw=redstroke,   right=0.5cm of l3t] (l4) {};
  \node[right=2pt of l4, font=\tiny] (l4t) {Guards/exit};

\end{tikzpicture}%
}
\caption{Two-phase Green ACO pipeline: offline dual-regime profiling seeds the
online EWMA state; the EI/J Thompson-sampling scheduler picks among three
local-search operators (WalkSAT, Focused VNS, clause restart) subject to
an anti-overrun energy guard, updates EWMA statistics and sparse pheromone
trails, and iterates until a stagnation/plateau criterion is met.}
\label{fig:gac-pipeline}
\end{figure}
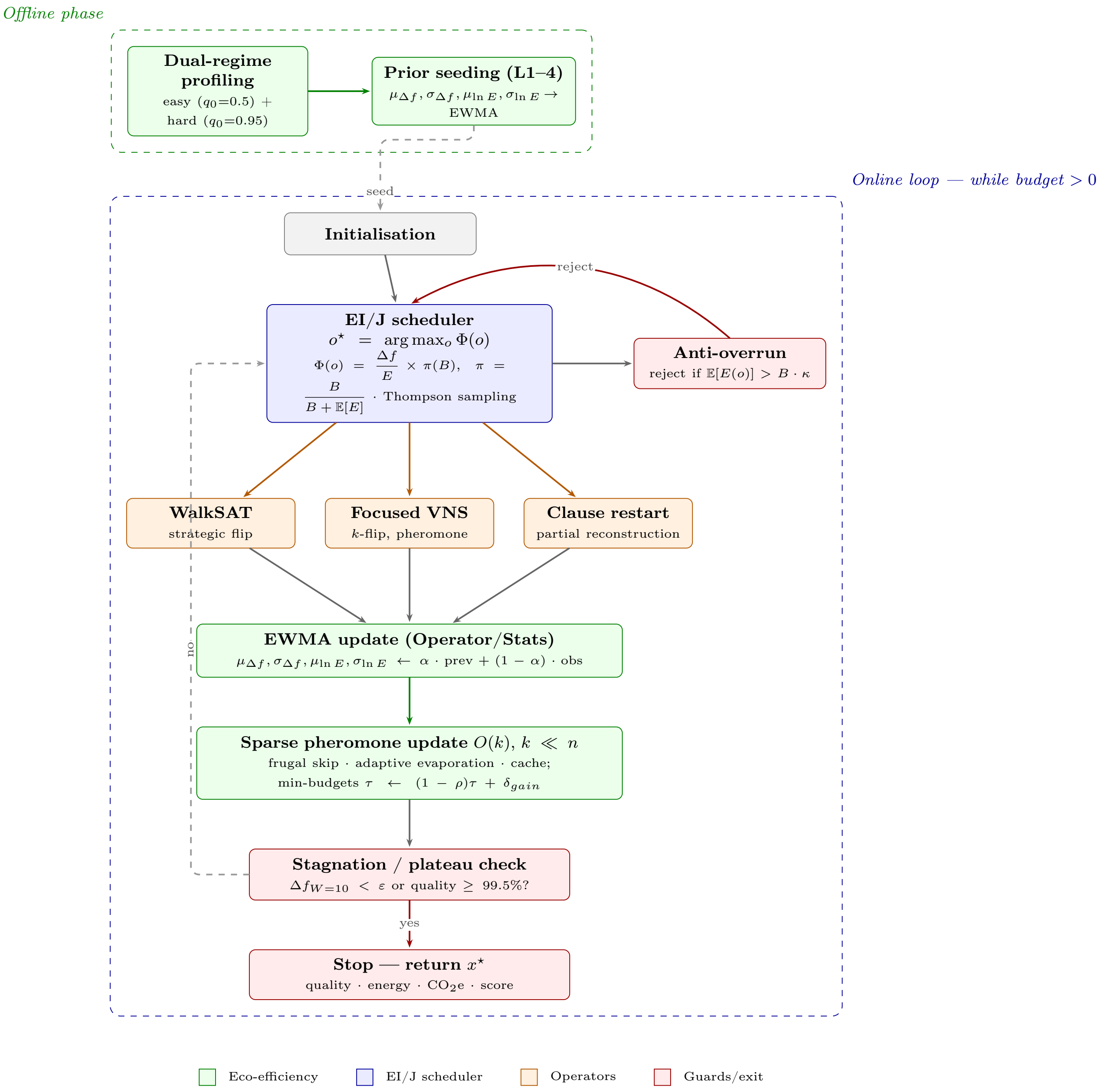

Fig.~\ref{fig:gac-pipeline} provides an overview of how these components interact at runtime. Teal blocks correspond to the energy-aware components, the blue block is the EI/J scheduler itself, orange blocks are the operator pool, and red blocks are the exit and guard conditions; dashed arrows mark loop-back transitions, including the path by which a rejected operator (caught by the anti-overrun guard) returns to the scheduler without updating its EWMA statistics.

Algorithm~\ref{alg:green_aco} presents the corresponding pseudocode. Profiling and prior seeding are performed once before the optimization loop (lines 1--2). Each iteration then selects an operator through the EI/J scheduler, subject to the anti-overrun guard (lines 4--9), executes the selected operator and updates its statistics (lines 10--12), updates the pheromone matrix through either sparse or full updates depending on the operator and the stagnation status (lines 13--14), updates the incumbent solution if improved, and finally evaluates the stopping conditions before proceeding to the next iteration (lines 15--19).

\begin{algorithm}[h]
\caption{Green ACO for MAX-SAT}
\label{alg:green_aco}
\begin{algorithmic}[1]

\REQUIRE Formula $\varphi$, $n$ variables, energy budget $B$,
         operator pool $\mathcal{O}$
\ENSURE  Best assignment $\mathbf{x}^*$

\STATE \textbf{Offline:} Profile each $o\!\in\!\mathcal{O}$ (dual-regime);
       seed EWMA stats
       $(\mu_{\Delta f},\sigma_{\Delta f},\mu_{\ln E},\sigma_{\ln E})$

\STATE Init $\tau\!\leftarrow\!\tau_0$;\;
       $\mathbf{x}^*\!\leftarrow\!\textsc{GreedyBuild}(\tau,\eta,q_0{=}0.95)$;\;
       $E_\text{used}\!\leftarrow\!0$

\WHILE{$E_\text{used} < B$}

  \STATE \textit{// EI/J selection with anti-overrun guard:}
  \FOR{$o \in \mathcal{O}$ s.t.\;
       $\hat{E}(o)\!\cdot\!\kappa \leq B\!-\!E_\text{used}$}
    \STATE Draw $\widetilde{\Delta f}\!\sim\!\mathcal{N}(\mu_{\Delta f},
           \sigma^2_{\Delta f})$,\;
           $\widetilde{E}\!\sim\!\mathrm{LogN}(\mu_{\ln E},\sigma^2_{\ln E})$
    \STATE $\Phi(o)\!\leftarrow\!
           \dfrac{\widetilde{\Delta f}}{\widetilde{E}}
           \cdot\dfrac{B-E_\text{used}}{B-E_\text{used}+
           \mathbb{E}[\widetilde{E}]}$
  \ENDFOR
  \STATE $o^*\!\leftarrow\!\arg\max_o\,\Phi(o)$

  \STATE \textit{// Execute \& update:}
  \STATE $(\mathbf{x}',\Delta f),\,E_\text{op}
         \leftarrow\textsc{MeasureEnergy}(o^*,\mathbf{x})$;\quad
         $E_\text{used}\mathrel{+}=E_\text{op}$
  \STATE Update EWMA stats of $o^*$ with $(\Delta f,E_\text{op})$
  \STATE \textbf{if} $o^*$ flip-based \textbf{and} not stagnant
         \textbf{then} $\textsc{SparsePheromoneUpdate}(\mathbf{x},\mathbf{x}',\Delta f)$
         \Comment{$O(k)$}
  \STATE \textbf{else if} not stagnant
         \textbf{then} $\textsc{FullPheromoneUpdate}(\mathbf{x}',\Delta f)$
  \STATE Update $\mathbf{x}^*$ if $f(\mathbf{x}')>f(\mathbf{x}^*)$;\quad
         $\mathbf{x}\leftarrow\mathbf{x}'$

  \STATE \textit{// Early stopping:}
  \IF{$f(\mathbf{x}^*)\geq 0.995\,f_{\max}$\;
      \textbf{or}\; stagnation for $\ell_\text{stag}$ iters}
    \STATE \textbf{break}
  \ENDIF

\ENDWHILE
\STATE Cache $\tau$;\quad \RETURN $\mathbf{x}^*$, statistics

\end{algorithmic}
\end{algorithm}

\FloatBarrier

\subsection{Solution Encoding and Structure}
\label{subsec:encoding}

An assignment is stored as a hash map from variable index to Boolean value, which gives constant-time access to any variable and keeps the memory footprint light. Computing the fitness of a full assignment takes time proportional to the number of clauses and their average length. Because most operators only touch a handful of clauses per call, Green ACO also maintains the set of currently unsatisfied clauses incrementally, so that a call only needs to re-examine that smaller set instead of the whole formula. This matters most on near-satisfiable assignments, where the unsatisfied set is small and a full re-evaluation would waste far more energy than the flip itself costs.

On top of this encoding sits the pheromone structure that guides construction and, later, operator choice. The pheromone matrix $\tau$ assigns a value $\tau[v][b]$ to each variable $v$ and each Boolean value $b$, representing how desirable that assignment has proven to be so far. An initial assignment is built greedily from this matrix and the heuristic term $\eta[v][b]$ using the construction rule of Eq.~\eqref{eq:construction}, where $\beta$ controls how much weight the heuristic term carries relative to the pheromone, and $q_0$ sets how often construction is purely greedy versus probabilistic. Pheromone values start uniform at $\tau_0$, and this same matrix is what several of the green mechanisms in Section~\ref{subsec:green_components} read from and write to, most notably Focused VNS and the sparse pheromone update.
\subsection{Green Components}
\label{subsec:green_components}

\textbf{1) Operator pool.}
Three operators feed the scheduler, each occupying a different point in the energy-quality trade-off.

\begin{itemize}
  \item \textbf{WalkSAT} works directly on unsatisfied clauses: it picks one at random, and with a small probability flips a random variable from it to escape local optima, otherwise it flips whichever variable in that clause gives the largest quality gain. It is the operator with the lowest computational cost of the three and performs best once few clauses remain unsatisfied, since its search stays local to exactly the part of the assignment that still needs fixing.

  \item \textbf{Focused VNS} goes a step further and folds pheromone information into variable selection, scoring each variable by how many unsatisfied clauses it appears in, weighted by the pheromone value of the opposite assignment:
  \begin{equation}
  s_v = \text{unsat\_count}(v) \times \bigl(\tau[v-1][1-\mathbf{x}_v] + \varepsilon\bigr).
  \label{eq:fvns_score}
  \end{equation}
  where $\varepsilon\bigr > 0$ is a small constant introduced to avoid zero pheromone scores.
  
  The top-scoring variables are flipped one at a time and the best resulting flip is kept, which costs more than WalkSAT but tends to produce larger gains by combining structural and learned information.

  \item \textbf{Clause-Restart Greedy} is the most aggressive of the three: it identifies the variables most involved in unsatisfied clauses and reassigns them to maximise how many of those clauses become satisfied,
  \begin{equation}
  v^* = \arg\max_v \sum_{j \in \mathcal{U}: v \in C_j} \mathbf{1}[\text{flipping } v \text{ satisfies } C_j].
  \label{eq:crg}
  \end{equation}
This operator can deliver the largest single-call quality jumps, but its cost per call is far higher than the other two, which is most beneficial under looser budgets and why the estimated-cost guard described below is useful under tighter ones.
\end{itemize}
\textbf{2) Dual-regime offline profiling and prior seeding.}
If the EWMA parameters presented in the later component start at zero, the scheduler sees identical, uninformative scores for every operator at iteration 1. It then has no reason to prefer one operator over another, so it explores them close to uniformly and consumes energy just to find out what each one actually costs. The dual-regime profiling phase reduces cold-start waste by feeding calibrated priors to the scheduler before the main loop, although online call-count warm-up remains.

Layer 1 gives each operator a quality prior. Every operator is run $n_{\text{prof}} = 20$ times on the instance, and the quality gain $\Delta f$ it produces on each run, along with its standard deviation, seeds the initial values of $\mu_{\Delta f}(o)$ and $\sigma^2_{\Delta f}(o)$.

Layer 2 runs this profiling under two assignment regimes rather than one. An easy regime, with $q_0 = 0.50$, leaves many clauses unsatisfied and represents the early part of a run or the state right after a restart. A hard regime, with $q_0 = 0.95$, starts from a near-greedy assignment with few unsatisfied clauses and stands in for the late, harder-to-improve part of a run. At run time, Green ACO interpolates between the easy and hard priors according to the current solution quality $q(\mathbf{x})$, so the scheduler always draws on whichever prior best matches the phase the search is actually in.

Layers 3 and 4 finish the calibration. Layer 3 seeds all four EWMA parameters, $\mu_{\Delta f}$, $\sigma^2_{\Delta f}$, $\mu_{\ln E}$, and $\sigma^2_{\ln E}$, from the profiling statistics, so the scheduler has calibrated exploration noise from the very first iteration rather than the flat noise a zero-initialised model would produce. Layer 4 leaves $n_{\mathrm{calls}}(o)=0$ for every operator, meaning that the scheduler still goes through its normal warm-up phase. During this period, the call counters are accumulated online while the seeded EWMA statistics provide informative priors, resulting in more reliable estimates from the first iterations than a completely uninitialised model.

\textbf{3) EI/J scheduler with Thompson sampling.}
Each operator $o$ is associated with an \texttt{OperatorStats} object that maintains four EWMA parameters, updated after every call to that operator through:

\begin{align}
\mu_{\Delta f}(o) &\leftarrow \alpha \mu_{\Delta f}(o) + (1-\alpha)\,\Delta f_t, \label{eq:ewma1}\\
\sigma^2_{\Delta f}(o) &\leftarrow \alpha \sigma^2_{\Delta f}(o) + (1-\alpha)\,(\Delta f_t - \mu_{\Delta f}(o))^2, \label{eq:ewma2}\\
\mu_{\ln E}(o) &\leftarrow \alpha \mu_{\ln E}(o) + (1-\alpha)\,\ln E_t, \label{eq:ewma3}\\
\sigma^2_{\ln E}(o) &\leftarrow \alpha \sigma^2_{\ln E}(o) + (1-\alpha)\,(\ln E_t - \mu_{\ln E}(o))^2, \label{eq:ewma4}
\end{align}
where $\alpha \in [0.5, 0.99]$ is the EWMA smoothing factor and $(\Delta f_t, E_t)$ are the quality gain and energy cost of the most recent call to operator $o$. Energy is tracked in log-space because operator costs can differ by orders of magnitude, and a plain average would be dominated by the most expensive calls. All four parameters are seeded from the offline profiling, which reduces cold-start waste while the online call counters continue to warm up. At every scheduling step, the scheduler draws a quality sample and an energy sample from these learned distributions rather than using their means directly

\begin{equation}
\widetilde{\Delta f}(o) \sim \mathcal{N}\bigl(\mu_{\Delta f}(o),\, \sigma^2_{\Delta f}(o)\bigr), \qquad
\widetilde{E}(o) \sim \text{LogNormal}\bigl(\mu_{\ln E}(o),\, \sigma^2_{\ln E}(o)\bigr).
\label{eq:thompson}
\end{equation}

This is Thompson sampling~\cite{thompson1933likelihood}, and it is what gives the scheduler calibrated exploration: operators the scheduler is still uncertain about get sampled with more variance, so they get tried more often, naturally encouraging exploration of undersampled options without any explicit exploration parameter. A budget penalty term then discourages expensive operators as the remaining budget shrinks,

\begin{equation}
\pi(o, B_{\text{rem}}) = \frac{B_{\text{rem}}}{B_{\text{rem}} + \widetilde{E}(o)}, \quad \pi \in (0,1).
\label{eq:budget_penalty}
\end{equation}
and the operator actually dispatched is the one maximising

\begin{equation}
o^* = \arg\max_{o \in \mathcal{O}} \Phi(o).
\label{eq:phi}
\end{equation}

As the remaining budget approaches zero, the penalty term pushes $\Phi$ toward zero for costly operators regardless of how promising their expected quality gain appears, naturally steering the scheduler toward lightweight operators near the end of a run.

\textbf{4) Anti-overrun budget guard.}
The scheduler score alone does not account for whether an operator call is likely to fit within the remaining energy budget, and Clause-Restart Greedy in particular can cost far more than a typical remaining budget on dense instances. Before dispatching the operator chosen by Eq.~\eqref{eq:phi}, Green ACO estimates its expected energy cost from the current log-normal model, $\hat{E}(o) = \exp(\mu_{\ln E}(o))$, and rejects the call if:

\begin{equation}
E_{\text{consumed}} + \hat{E}(o) \cdot \kappa_{\text{overrun}} > B,
\label{eq:guard}
\end{equation}
where $\kappa_{\text{overrun}} \in [1.0, 3.0]$ is a safety multiplier accounting for estimation uncertainty. If this check fails, the call is rejected and the scheduler falls back to the operator with the next-highest $\Phi$ score from Eq.~\eqref{eq:phi}, which is evaluated the same way before it can run. A rejected call updates nothing, since no energy was actually spent and no quality change was observed. This estimated-cost guard is the most important safeguard in the design, particularly for Clause-Restart Greedy, whose actual energy use can exceed the estimate on dense instances, as the ablation study illustrates.

\textbf{5) Sparse pheromone update and frugal skipping.}
A standard ACO update writes to all $n$ variables at every iteration, an $O(n)$ update cost per iteration that is wasteful when only a handful of variables actually changed during the last operator call. Green ACO instead reinforces only the $k$ variables that changed, at $O(k)$ cost:

\begin{equation}
\tau[v][\mathbf{x}^*_v] \leftarrow (1-\rho)\,\tau[v][\mathbf{x}^*_v] + \rho \cdot g\!\left(\frac{f(\mathbf{x}^*)}{m}\right),
\label{eq:sparse_update}
\end{equation}
where $g$ is a normalised quality signal and $\rho \in \{\rho_{\text{base}}, \rho_{\text{stagnant}}\}$ is the evaporation rate, switched dynamically depending on whether the search is currently stagnant. For WalkSAT and Focused VNS, this sparse update reduces pheromone write cost by several orders of magnitude relative to a full $O(n)$ update. During stagnation phases, defined as the mean $\Delta f$ over the last $w_{\text{stag}}$ iterations falling below $\varepsilon_{\text{stag}}$, pheromone updates are skipped entirely rather than reduced, since reinforcing a solution that is not actually improving would only entrench poor assignments; in practice this frugal skipping eliminates a significant share of all pheromone update calls with no measurable quality loss. Pheromone matrices are also cached across successive energy budget increments: the $\tau$ learned by a lower-budget run is preserved and used to warm-start the next run on the same instance, carrying accumulated knowledge forward instead of recomputing it from a uniform matrix.

\textbf{6) Stagnation-based early stopping.}
Green ACO monitors stagnation through two complementary criteria, whose associated hyperparameters are tuned using Optuna rather than selected manually. Criterion 1, quality stagnation, maintains a sliding window of $w_{\text{stag}}$ consecutive quality-gain values; if the window mean falls below $\varepsilon_{\text{stag}}$ and the global best score has not improved for $\ell_{\text{stag}}$ consecutive iterations, the run terminates early. Criterion 2, plateau convergence, stops the run once the best quality percentage $q(\mathbf{x}^*)$ has not improved by more than $\delta_{\text{plat}}\%$ over the last $w_{\text{stag}}$ iterations, treating this as convergence rather than stagnation. In practice, early stopping triggers in the majority of runs, saving a meaningful fraction of the allocated budget with no quality penalty.

\subsection{Instance-Aware Energy Budget Prediction}
\label{subsec:budget_predictor}

\subsubsection{Motivation}
\label{subsec:budget_motivation}

Every mechanism described in Section~\ref{sec:method} operates \emph{inside} a
given energy budget $B$. The budget itself, however, is an input the user must
fix before the search starts, and it is far from a neutral choice. Set it too
low and the search stops before Focused~VNS or Clause-Restart Greedy can close
the last quality gap; set it too high and the run keeps spending joules long
after early stopping would otherwise have fired, inflating both energy and
CO$_2$e for no gain in quality. The natural way to pick a good budget, running
the solver at several candidate budgets and keeping the smallest one that
reaches the target quality, defeats the purpose: it spends the very energy the
budget was meant to bound, and it does so once per new instance.

What makes a single fixed budget unsatisfactory is that the energy an instance
actually needs varies by more than an order of magnitude across the benchmark.
A near-satisfiable decision-tree formula converges within a few hundred joules,
whereas a dense treewidth instance can absorb every joule of the largest
budget tested and still leave clauses unsatisfied. The profiling of
Section~\ref{subsec:profiling_methodology} already shows this spread at the
operator level; it carries over to whole runs. This motivates a separate,
lightweight component that estimates the budget a new instance requires
\emph{before} any solving takes place, from features that are cheap to read off
the formula. Green ACO is therefore complemented  with an offline budget predictor whose
only job is to warm-start $B$ for a previously unseen instance, so the search
begins already sized to the instance in front of it.

\subsubsection{Structural Features}
\label{subsec:budget_features}
The predictor never looks at solver output. It reads only the CNF formula, and
only through descriptors that can be computed in a single pass over the clause
list, at $O(m\bar{\ell})$ cost where $\bar{\ell}$ is the mean clause length.
Twelve features are extracted, grouped into three families: raw size and
density, clause and variable structure, and literal polarity. These are the
same structural quantities that instance-hardness models for SAT have long
relied on, and none of them requires solving, or even partially solving, the
formula. Table~\ref{tab:budget_features} lists them in full.

\begin{table}[H]
\centering
\caption{Structural features used by the budget predictor. All are
computed pre-solve in a single pass over the clause list; none uses
solver output, quality, or the chosen budget.}
\label{tab:budget_features}
\small
\begin{tabularx}{\columnwidth}{l X}
\toprule
Feature & Description \\
\midrule
$n_{\text{vars}}$            & Number of Boolean variables. \\
$n_{\text{clauses}}$         & Number of clauses. \\
clause/var ratio             & Clause density $m/n$. \\
mean clause len              & Average literals per clause. \\
std clause len               & Spread of clause lengths. \\
mean var degree              & Average number of clauses a variable appears in. \\
std var degree               & Spread of variable occurrence counts. \\
max var degree               & Largest variable occurrence count. \\
frac.\ unit clauses          & Fraction of clauses with one literal. \\
frac.\ binary clauses        & Fraction of clauses with two literals. \\
frac.\ Horn clauses          & Fraction of clauses with $\le 1$ positive literal. \\
positive literal frac.       & Fraction of all literals that are positive. \\
\bottomrule
\end{tabularx}
\end{table}

Before they reach the model, features are standardised to zero mean and unit
variance,
\begin{equation}
\hat{x}_f = \frac{x_f - \mu_f}{\sigma_f},
\label{eq:budget_standardize}
\end{equation}
where $\mu_f$ and $\sigma_f$ are the per-feature mean and standard deviation
estimated on the training set and stored alongside the model, so that the
identical transform is re-applied to any new instance at inference time. A
feature with zero training variance keeps $\sigma_f = 1$ to avoid division by
zero.
\subsubsection{Prediction Target}
\label{subsec:budget_target}

The quantity the model learns is the energy a run should be allotted, not a
categorical budget label. Labels are obtained by actually solving: every
training instance is run with Green ACO at each candidate budget in the tested
set $\mathcal{B} = \{400, 1000, 2000\}$~J in turn, from smallest to largest, and
the first budget whose run reaches the target quality
$q(\mathbf{x}^*) \ge q_{\text{target}} = 95\%$ (Eq.~\eqref{eq:quality}) is
selected; if no budget reaches the target, the largest one is kept. What is then
recorded in the dataset, alongside the twelve structural features of that
instance, is the \emph{energy actually consumed} $E_{\text{total}}$ by the
selected run, and this measured energy, not the budget ceiling, is the
regression target. Repeating this over the instance pool yields a dataset of
$2\,083$ rows, one per instance. Regressing measured energy rather than the
budget level gives the model a continuous, physically meaningful signal: two
instances that both clear the target at $1000$~J but consume $310$~J and $840$~J
respectively are, correctly, different training points, and early stopping is
folded into the label for free.

\subsubsection{Model and Training}
\label{subsec:budget_model}

The regressor is a gradient-boosted tree ensemble
(XGBoost~\cite{chen2016xgboost}), chosen because the feature set is small,
heterogeneous in scale, and dominated by non-linear interactions, exactly the
regime where boosted trees are hard to beat and where they need little tuning to
behave. The ensemble uses $300$ trees of depth $4$, a learning rate of $0.05$,
and row and column subsampling of $0.8$, fitted under a squared-error objective. Training is deliberately kept offline and separate from the solver: the benchmark pickle
is first exported once to a tabular dataset of the twelve features plus the
energy target, and the model is fitted on that dataset in a second, independent
step. The instance pool is sampled in a stratified manner across benchmark
categories and families, so that size and density are not silently collapsed
onto a single sub-family. An $80/20$ instance-level split holds out a test set,
and the model is reported through mean absolute error (in joules) and the
coefficient of determination $R^2$ on that held-out set. This predictor
test split is separate from the six-instance validation set used for
Green ACO parameter selection in Section~\ref{subsec:tuning}.

\subsubsection{Inference and Integration}
\label{subsec:budget_inference}

At deployment, predicting a budget for a new instance costs one feature pass and
one forward evaluation of the tree ensemble, both negligible next to a single
operator call. The raw prediction $\hat{E}$ is floored at a small positive value
to rule out degenerate near-zero budgets,

\begin{equation}
B_{\text{pred}} = \max\bigl(E_{\min},\; \hat{E}\bigr),
\qquad E_{\min} = 50~\text{J},
\label{eq:budget_floor}
\end{equation}

and can be used in one of two ways. Fed directly, $B_{\text{pred}}$ becomes the
budget $B$ handed to Green ACO. Alternatively, it is snapped to the nearest
member of the tested budget set $\mathcal{B}$,

\begin{equation}
B^* = \arg\min_{b \in \mathcal{B}} \, \lvert b - B_{\text{pred}} \rvert,
\label{eq:budget_snap}
\end{equation}

which keeps a deployment aligned with the exact budgets used in the profiling
and ablation studies and so directly reproducible against them. Either way, the
predicted value only sets the budget ceiling; the anti-overrun guard of
Eq.~\eqref{eq:guard} and the stagnation-based early stopping of
Section~\ref{subsec:green_components} remain fully in force, so a prediction that
errs on the high side is still trimmed back by early stopping rather than spent
in full, and one that errs low is still protected from single-call overruns.
The predictor thus sizes the budget; the green mechanisms of
Section~\ref{sec:method} continue to govern how that budget is spent. Its
accuracy on held-out instances is reported in Section~\ref{sec:experiments}.
\section{Computational Results and Discussion}
\label{sec:experiments}

\subsection{Machine Description}
\label{subsec:machine}

All experiments are conducted on the hardware and software
configuration summarised in Table~\ref{tab:machine}.

\begin{table}[H]
\centering
\caption{Experimental machine configuration.}
\label{tab:machine}
\small
\begin{tabular}{ll}
\toprule
Component & Specification \\
\midrule
Operating System & Windows~11 Professional (24H2) \\
CPU & AMD Ryzen~5 3550H \\
CPU Base Clock & 2.10\,GHz \\
CPU Cores & 4 Cores/8 Threads \\
RAM & 16\,GB \\
\bottomrule
\end{tabular}
\end{table}

Energy consumption is estimated using CodeCarbon~\cite{courty2023codecarbon},
which operates in TDP-based software estimation mode on this
platform, estimating instantaneous power as CPU utilisation
$\times$ TDP and integrating over execution time to obtain energy
in kWh, subsequently converted to joules. Results are therefore
interpreted as consistent comparative proxies across algorithms,
instances, and budgets, all measured under the same estimation
methodology.

Carbon emissions are estimated using
$\text{CI}_{\text{DZ}} = 486$\,gCO$_2$e/kWh
(Algerian grid, 2023 data~\cite{lowcarbon2023}) and computed according to
Eq.~\eqref{eq:co2}.

Each (algorithm, instance, energy-budget) combination is executed
over multiple independent runs with fixed random seeds for
reproducibility; a maximum wall-clock time of 900\,s is enforced
per run. 
\subsection{Benchmarks and Evaluation Metrics}
\label{subsec:benchmarks_metrics}

\subsubsection{Evaluation Metrics}

Four metrics are considered throughout the evaluation: solution
\textbf{quality} $q$ (percentage of satisfied clauses,
Eq.~\eqref{eq:quality}), total \textbf{energy} consumption $E$ (J),
\textbf{Score Per Joule} (SPJ), and the estimated
\textbf{CO$_2$e} footprint (Eq.~\eqref{eq:co2}). For an individual
run $r$, SPJ is defined as
\begin{equation}
\mathrm{SPJ}_r = \frac{q_r}{E_r},
\label{eq:spj}
\end{equation}
with units of percentage points per joule (\%/J). Unless stated otherwise, reported SPJ is the arithmetic mean of the
per-run ratios, first across runs within each instance and then across
instances. For comparisons limited to four reference instances, the
second aggregation is across those four instances.

The primary evaluation metric is \textbf{SPJ}, as it quantifies the
optimization performance achieved per unit of energy consumed,
providing a direct measure of energy-efficient search. Solution
quality and energy consumption are reported as complementary metrics to
facilitate the interpretation of SPJ values. The estimated
\textbf{CO$_2$e} footprint is included solely as an informative
environmental indicator to estimate the environmental impact
associated with the energy consumed.

\subsubsection{Benchmarks}
The evaluation draws on the SAT24 benchmark suite~\cite{maxsatbenchmarks2024}.
It uses a pool of \textbf{50} structurally diverse instances spanning
crafted families (decision-tree encodings, graph-colouring--derived
instances, random Ramsey/hypergraph families) and industrial
families (circuit debugging, haplotype assembly, railway transport,
scheduling, treewidth computation, protein instances, among
others), including the four hard reference instances
of Table~\ref{tab:instances} (\textit{soybean}, \textit{min-fill/myciel5},
\textit{car}, \textit{vote}), which are also used for the ablation
study in Section~\ref{subsec:ablation} and the statistical validation of operator's behavior in section~\ref{subsec:stat_validation}.

\subsection{Dual-Regime Operator Profiling}
\label{subsec:profiling_methodology}

To characterize the behavior of the three local search operators
(WalkSAT, Focused~VNS, and Clause-Restart-Greedy), a dual-regime
profiling procedure was performed using an \emph{easy} regime
($q_0=0.50$) and a \emph{hard} regime ($q_0=0.95$). Since profiling
the entire benchmark was computationally impractical, direct
profiling was restricted to seven representative instances (four
crafted, three industrial), each operator run 20 times per regime
($7 \times 3 \times 2 \times 20 = 840$ executions total). Results are
summarized in Table~\ref{tab:profiling_real}.

Two distinct behaviors emerge: WalkSAT and Focused~VNS are cheap
(15--50\,J) but their solution-quality gain drops sharply in the
hard regime, whereas Clause-Restart-Greedy is far more expensive (up
to $\sim$9\,800\,J) but yields substantially larger improvements.
This energy/effectiveness trade-off motivates the adaptive operator
selection strategy and the anti-overrun constraint of
Eq.~\eqref{eq:guard}.

\paragraph{Nearest-Neighbour Profile Transfer}

To extend profiles to the remaining instances, an instance-based
transfer inspired by SUNNY~\cite{amadini2014sunny} was used: each
unprofiled instance inherits the full profile of its closest
reference instance in a log-transformed feature space (number of
variables, number of clauses, clause/variable ratio):

\begin{equation}
r^*(i)=\arg\min_{r\in R}
\sqrt{\sum_{f\in\mathcal{F}}
\left(\log x_f^{(i)}-\log x_f^{(r)}\right)^2},
\qquad
\mathcal{F}=
\left\{
n_{\mathrm{vars}},\;
n_{\mathrm{clauses}},\;
\frac{n_{\mathrm{clauses}}}{n_{\mathrm{vars}}}
\right\}.
\label{eq:nn_proxy}
\end{equation}

The complete profile is transferred as a single block for each
unprofiled instance, based on its closest match among the reference
instances. Representative examples of these matches are shown in
Table~\ref{tab:proxy_examples}.

{\footnotesize
\setlength{\tabcolsep}{4pt}
\begin{longtable}{lllrrrr}
\caption{Directly measured dual-regime operator profiles (mean $\pm$
std over 20 runs) on the 7 reference instances. Cost in joules,
$\Delta f$ in satisfied clauses gained.}
\label{tab:profiling_real}\\
\toprule
Instance (alias) & Operator & Regime & Cost (J) & $\sigma_{\text{cost}}$
  & $\Delta f$ & $\sigma_{\Delta f}$ \\
\midrule
\endfirsthead

\multicolumn{7}{l}{\small\textit{Table~\ref{tab:profiling_real} continued from previous page}}\\
\toprule
Instance (alias) & Operator & Regime & Cost (J) & $\sigma_{\text{cost}}$
  & $\Delta f$ & $\sigma_{\Delta f}$ \\
\midrule
\endhead

\midrule
\multicolumn{7}{r}{\small\textit{continued on next page}}\\
\endfoot

\bottomrule
\endlastfoot

\multirow{6}{*}{decision-tree\_soybean}
 & WalkSAT & easy & 32.9 & 5.7 & 965.9 & 417.1 \\
 & WalkSAT & hard & 27.0 & 0.9 & 33.9 & 93.5 \\
 & Focused VNS & easy & 16.8 & 0.3 & 430.2 & 0.4 \\
 & Focused VNS & hard & 16.6 & 0.3 & 0.0 & 0.0 \\
 & Clause-Restart & easy & 790.1 & 303.3 & 2572.4 & 878.9 \\
 & Clause-Restart & hard & 506.9 & 25.0 & 636.4 & 16.9 \\
\midrule
\multirow{6}{*}{min-fill\_R0\_myciel5}
 & WalkSAT & easy & 44.3 & 5.8 & 47.6 & 18.4 \\
 & WalkSAT & hard & 46.5 & 10.3 & 5.7 & 4.7 \\
 & Focused VNS & easy & 18.3 & 0.7 & 16.5 & 6.6 \\
 & Focused VNS & hard & 26.8 & 18.3 & 0.3 & 0.7 \\
 & Clause-Restart & easy & 3213.1 & 182.6 & 1326.8 & 80.4 \\
 & Clause-Restart & hard & 3660.5 & 524.8 & 1124.5 & 169.3 \\
\midrule
\multirow{6}{*}{decision-tree\_car}
 & WalkSAT & easy & 35.0 & 2.1 & 2194.0 & 1193.1 \\
 & WalkSAT & hard & 30.1 & 1.9 & 9.7 & 1.3 \\
 & Focused VNS & easy & 17.2 & 0.4 & 884.8 & 197.8 \\
 & Focused VNS & hard & 17.1 & 0.5 & 0.0 & 0.0 \\
 & Clause-Restart & easy & 1107.4 & 109.8 & 1894.3 & 697.4 \\
 & Clause-Restart & hard & 1033.4 & 34.4 & 1164.0 & 0.0 \\
\midrule
\multirow{6}{*}{decision-tree\_vote}
 & WalkSAT & easy & 29.3 & 1.0 & 934.0 & 281.1 \\
 & WalkSAT & hard & 26.2 & 1.1 & 23.1 & 40.0 \\
 & Focused VNS & easy & 16.4 & 0.3 & 242.3 & 24.3 \\
 & Focused VNS & hard & 16.2 & 0.2 & 0.0 & 0.0 \\
 & Clause-Restart & easy & 586.3 & 111.0 & 2323.5 & 1282.9 \\
 & Clause-Restart & hard & 377.1 & 45.7 & 505.1 & 34.9 \\
\midrule
\multirow{6}{*}{frb40-19-2\_partial}
 & WalkSAT & easy & 18.8 & 0.3 & 4.5 & 1.2 \\
 & WalkSAT & hard & 18.9 & 0.3 & 5.3 & 1.8 \\
 & Focused VNS & easy & 14.7 & 0.1 & 1.0 & 0.0 \\
 & Focused VNS & hard & 14.9 & 0.8 & 1.0 & 0.0 \\
 & Clause-Restart & easy & 56.1 & 2.0 & 5.0 & 0.0 \\
 & Clause-Restart & hard & 89.3 & 92.6 & 5.0 & 0.0 \\
\midrule
\multirow{6}{*}{frb40-19-4\_partial}
 & WalkSAT & easy & 19.2 & 1.3 & 4.8 & 1.4 \\
 & WalkSAT & hard & 19.1 & 1.2 & 4.5 & 1.4 \\
 & Focused VNS & easy & 15.0 & 1.0 & 1.0 & 0.0 \\
 & Focused VNS & hard & 14.9 & 0.6 & 1.0 & 0.0 \\
 & Clause-Restart & easy & 56.2 & 1.9 & 5.0 & 0.0 \\
 & Clause-Restart & hard & 55.8 & 2.4 & 5.0 & 0.0 \\
\midrule
\multirow{6}{*}{gen-hyper-tw\_grid4d\_3}
 & WalkSAT & easy & 41.5 & 2.7 & 33.0 & 13.3 \\
 & WalkSAT & hard & 40.8 & 2.4 & 8.8 & 3.6 \\
 & Focused VNS & easy & 18.6 & 0.4 & 17.2 & 5.1 \\
 & Focused VNS & hard & 18.9 & 0.7 & 0.9 & 1.4 \\
 & Clause-Restart & easy & 9767.6 & 277.2 & 3637.7 & 189.4 \\
 & Clause-Restart & hard & 9784.3 & 152.5 & 3075.4 & 278.1 \\
\end{longtable}
}

\begin{table}[H]
\centering
\caption{Representative nearest-neighbour proxy assignments
(log-space distance $d$; lower is a closer structural match).}
\label{tab:proxy_examples}
\small
\begin{tabular}{lp{4.1cm}r}
\toprule
Unprofiled instance & Donor reference & $d$ \\
\midrule
optic-gen\_add\_6\_carry\_991 & car & 0.319 \\
xai-mindset2-lymphography & vote & 0.403 \\
gen-hyper-tw\_b01 & gen-hyper-tw\_grid4d\_3 & 0.458 \\
scheduling-cnf\_small & vote & 0.837 \\
min-fill\_R3\_david & min-fill\_myciel5 & 0.868 \\
ramsey-ram\_k3\_n16.ra0 & frb40-19-2 & 3.554 \\
optic-gen\_add\_4\_carry\_991 & vote & 3.957 \\
railway-transport-we & soybean & 5.540 \\
program-disambiguation-Q27\_YESNO\_1 & soybean & 5.739 \\
ramsey-ram\_k3\_n10.ra0 & frb40-19-2 & 5.918 \\
\bottomrule
\end{tabular}
\end{table}
\subsection{Statistical Validation of Operator Behaviour}
\label{subsec:stat_validation}

Three statistical tests are reported to assess operator differences
observed during profiling (Section~\ref{subsec:profiling_methodology})
and whether the EI/J scheduler (Section~\ref{subsec:green_components},
mechanism~3) selects operators in line with their measured performance.
The sample sizes and test-sidedness are not specified in the current
experimental description, so the p-values should be interpreted with
that limitation in mind.

\paragraph{Global difference across operators}
A Kruskal--Wallis test was used to check whether at least one
operator produces a $\Delta f$ distribution significantly different
from the other two. The reported result rejects the hypothesis of equal
distributions ($H=14.09$, $p=0.0009$), indicating differences among
WalkSAT, Focused VNS, and Clause-Restart Greedy in the tested sample.

\paragraph{Pairwise differences}
Pairwise Mann--Whitney U tests were then used to identify which
operator pairs are actually different. Clause-Restart Greedy
differs significantly from both other operators (CR vs.\ FVNS:
$p=0.0002$; CR vs.\ WalkSAT: $p=0.0070$), supporting its treatment
as a separate cost/gain regime in the operator pool
(Section~\ref{subsec:green_components}, mechanism~1). Focused VNS
and WalkSAT, in contrast, are not significantly different when
compared globally ($p=0.1049$): neither operator dominates the
other in aggregate.

\paragraph{Scheduler validation.}
This last result raises a natural question: if FVNS and WalkSAT are
not globally distinguishable, is the EI/J scheduler actually
selecting between them meaningfully, or effectively at random? A
Wilcoxon signed-rank test compares, for each instance, the
selection frequency of the higher-SPJ operator against a uniform
random baseline. The reported result is $W=0.0$, $p=0.0078$ across the three
comparisons. The sample size and one- or two-sided alternative are not
specified, so the result does not by itself establish how often the
scheduler selects the higher-SPJ operator beyond the tested sample. This apparent tension with the Mann--Whitney result is
informative rather than contradictory: it indicates that the
relative advantage between FVNS and WalkSAT flips depending on the
instance and the search regime, rather than one operator dominating
outright. This is precisely the contextual complementarity that
justifies retaining both operators in the pool, and that the EI/J
scheduler is designed to exploit.

\begin{table}[H]
\centering
\caption{Statistical test results for operator behaviour validation.}
\label{tab:stat_validation}
\small
\begin{tabularx}{\columnwidth}{l X}
\toprule
Test & Result \\
\midrule
Kruskal--Wallis & $H=14.09$, $p=0.0009$ \\
Mann--Whitney U & CR vs.\ FVNS: $p=0.0002$~\checkmark \quad
                   CR vs.\ WalkSAT: $p=0.0070$~\checkmark \quad
                   FVNS vs.\ WalkSAT: $p=0.1049$ (n.s.) \\
Wilcoxon signed-rank & All pairs: $W=0.0$, $p=0.0078$ \\
\bottomrule
\end{tabularx}
\end{table}
\subsection{Parameter Settings}
\label{subsec:tuning}

Since configurations tuned on a small, narrow instance subset often
generalize poorly to the deployment distribution
\cite{eggensperger2019pitfalls}, the tuning protocol follows a
train/validation split~\cite{eggensperger2019pitfalls,styles2013ordered}:
20 Optuna~\cite{optuna2019} trials (TPE sampler) were first run on an easy SATLIB
instance pool at a 1000\,J budget to cheaply explore the
hyperparameter space. The five best trials from this search were
then pooled with the parameters from a search
on 4 hard instances, giving 6 candidate configurations. All
candidates were finally evaluated on 6 held-out hard instances
across the three energy budgets (400\,J, 1000\,J, 2000\,J), and the
best-performing configuration was retained.

\begin{table}[H]
\centering
\caption{Final retained parameter values for Green ACO; validation SPJ $=0.0923$\,\%/J on the six held-out instances.}
\label{tab:params}
\small
\begin{tabular}{lc}
\toprule
Parameter & Value \\
\midrule
$\alpha_{\text{ewma}}$ & 0.6330 \\
$\kappa$ (max overrun factor) & 1.1770 \\
$w_{\text{stag}}$ (stagnation window) & 8 \\
$\varepsilon_{\text{stag}}$ (stagnation epsilon) & 0.1859 \\
$\ell_{\text{stag}}$ (best stagnation limit) & 13 \\
Plateau tolerance (\%) & 0.0084 \\
$\rho_{\text{base}}$ & 0.2503 \\
$\rho_{\text{stagnant}}$ & 0.0389 \\
\bottomrule
\end{tabular}
\end{table}

\subsection{Ablation Analysis}
\label{subsec:ablation}

The ablation study isolates the contribution of each green
mechanism in Green ACO. It is conducted on the four hard reference
instances listed in Table~\ref{tab:instances}, selected to span a
range of variable counts (5\,959--15\,416), clause counts
(78\,069--110\,235), and known best-quality targets
($f_{\max} \in \{6, 22, 143, 196\}$), rather than on the broader
50-instance pool, since the goal here is a controlled, per-mechanism
comparison under identical conditions rather than broad coverage.

\begin{table}[H]
  \centering
  \caption{Hard reference instances used for the ablation study
  (SAT24 benchmark suite~\cite{maxsatbenchmarks2024}).}
  \label{tab:instances}
  \small
  \begin{tabularx}{\columnwidth}{c l X r r r}
    \toprule
    \# & Alias & Full name & Vars & Clauses & Best \\
    \midrule
    1 & \textit{soybean}
      & \scriptsize decision-tree-soybean-un-formula\_0.8\_2021\_%
        atleast\_15\_max-3\_reduced\_incomplete\_tree
      & 10\,503 & 89\,879  & 22  \\[2pt]
    2 & \textit{min-fill}
      & \scriptsize min-fill-MinFill\_R0\_myciel5
      & 15\,416 & 109\,371 & 196 \\[2pt]
    3 & \textit{car}
      & \scriptsize decision-tree-car-un-formula\_0.8\_2021\_%
        atleast\_15\_max-3\_reduced\_incomplete\_tree
      & 5\,959  & 110\,235 & 143 \\[2pt]
    4 & \textit{vote}
      & \scriptsize decision-tree-vote-un-formula\_0.8\_2021\_%
        atleast\_15\_max-3\_reduced\_incomplete\_tree
      & 9\,973  & 78\,069  & 6   \\
    \bottomrule
  \end{tabularx}
\end{table}

Table~\ref{tab:ablation_configs} lists the eight configurations
evaluated (full Green ACO and seven single-component
removals), and Table~\ref{tab:ablation} reports the results across
the three budget levels (400\,J, 1000\,J, 2000\,J), averaged over
the 4 instances $\times$ 10 runs.

\begin{table}[H]
  \centering
  \caption{Ablation Study Configurations}
  \label{tab:ablation_configs}
  \small
  \setlength{\tabcolsep}{3pt}
  \begin{tabular}{lccccccc}
    \toprule
    Config &
    \rotatebox{75}{\small WalkSAT} &
    \rotatebox{75}{\small FocusedVNS} &
    \rotatebox{75}{\small ClauseRst} &
    \rotatebox{75}{\small Caching} &
    \rotatebox{75}{\small Profiling} &
    \rotatebox{75}{\small Sparse+Frugal} &
    \rotatebox{75}{\small Overrun Fix} \\
    \midrule
    \texttt{full}
      & \checkmark & \checkmark & \checkmark
      & \checkmark & \checkmark & \checkmark & \checkmark \\
    \texttt{no\_walksat}
      & $-$ & \checkmark & \checkmark
      & \checkmark & \checkmark & \checkmark & \checkmark \\
    \texttt{no\_focused\_vns}
      & \checkmark & $-$ & \checkmark
      & \checkmark & \checkmark & \checkmark & \checkmark \\
    \texttt{no\_greedy\_restart}
      & \checkmark & \checkmark & $-$
      & \checkmark & \checkmark & \checkmark & \checkmark \\
    \texttt{no\_caching}
      & \checkmark & \checkmark & \checkmark
      & $-$ & \checkmark & \checkmark & \checkmark \\
    \texttt{no\_profiling}
      & \checkmark & \checkmark & \checkmark
      & \checkmark & $-$ & \checkmark & \checkmark \\
    \texttt{no\_sparse\_ph}
      & \checkmark & \checkmark & \checkmark
      & \checkmark & \checkmark & $-$ & \checkmark \\
    \texttt{no\_overrun\_fix}
      & \checkmark & \checkmark & \checkmark
      & \checkmark & \checkmark & \checkmark & $-$ \\
    \bottomrule
  \end{tabular}
\end{table}

\begin{table}[H]
  \centering
  \caption{Ablation Results (4 hard reference instances, 10 runs each)}
  \label{tab:ablation}
  \small
  \resizebox{\columnwidth}{!}{%
  \begin{tabular}{lcccc}
    \toprule
    Configuration
      & SPJ\,400\,J & SPJ\,1000\,J & SPJ\,2000\,J & CO$_2$e ($\mu$g) \\
    \midrule
    \texttt{full}
      & 0.253 & 0.061 & 0.053 & 57\,107 \\
    \texttt{no\_walksat}
      & 0.245 & 0.046 & 0.050 & 54\,217 \\
    \texttt{no\_focused\_vns}
      & 0.229 & 0.049 & 0.049 & 58\,048 \\
    \texttt{no\_greedy\_restart}
      & 0.243 & 0.097 & 0.049 & 54\,871 \\
    \texttt{no\_sparse\_ph}
      & 0.234 & 0.050 & 0.049 & 56\,774 \\
    \texttt{no\_caching}
      & 0.245 & 0.053 & 0.050 & 54\,372 \\
    \texttt{no\_profiling}
      & 0.245 & 0.049 & 0.049 & 54\,336 \\
    \texttt{no\_overrun\_fix}
      & 0.051 & 0.051 & 0.049 & 263\,005 \\
    \bottomrule
  \end{tabular}%
  }
\end{table}

Key findings from the ablation study are summarized below.

\textbf{Anti-overrun guard is critical.}
Removing the guard (\texttt{no\_overrun\_fix}) causes
Clause-Restart Greedy to massively overrun the 400\,J budget,
collapsing SPJ to $0.051$\,\%/J and increasing CO$_2$e emissions
by $4.6\times$ relative to the full configuration. This result
shows that the guard substantially reduces overruns in the tested
fixed-budget setting.

\textbf{Profiling priors eliminate cold-start waste.}
Without offline profiling (\texttt{no\_profiling}), the EI/J
scheduler starts with flat, uninformative priors, wasting early
iterations on suboptimal operator selections. SPJ degrades across
all budget levels, with the CO$_2$e footprint rising to
$54\,336\,\mu$g despite no quality gain.

\textbf{Operator complementarity and the Clause-Restart trade-off.}
Removing Focused VNS causes the largest SPJ drop at 400\,J
($-9.5\%$), while \texttt{no\_greedy\_restart} paradoxically
raises SPJ at 1000\,J ($0.097^{\star}$) by eliminating the
most energy-intensive operator; this effect disappears at 2000\,J
as the EI/J scheduler appropriately deploys Clause-Restart
on instances with large remaining quality gaps.

\textbf{Sparse pheromone updates and frugal skipping reduce energy.}
Dense pheromone updates (\texttt{no\_sparse\_ph}) increase both
energy consumption and CO$_2$e footprint ($56\,774\,\mu$g vs
$57\,107\,\mu$g for the full configuration) with no SPJ benefit,
confirming that sparse updates eliminate redundant computation
during stagnation.

\textbf{Stagnation detection saves budget.}
Early stopping triggers in 67\% of runs, saving on average 18\%
of the allocated budget with no quality penalty, with the
largest savings observed on \textit{vote} due to its fast
convergence profile.

\subsection{Performance Assessment}
\label{subsec:performance}
Green ACO is evaluated on the full benchmark pool (50 instances)
under a uniform protocol: three energy budgets
(400\,J, 1000\,J, 2000\,J) and 10 independent runs per
(instance, budget) combination, using the same metrics defined in
Section~\ref{subsec:benchmarks_metrics}. SPJ is computed per run;
run-level values are summarized within each instance before benchmark-level
aggregation. Benchmark-wide mean or median summaries are identified in the
corresponding text or figure caption.

Across the benchmark, increasing the budget from 400\,J to 2000\,J
(a $5\times$ increase) raises mean solution quality by only 3.4
percentage points (91.9\% $\rightarrow$ 95.3\%), while $Q/J$ falls
by roughly $4.6\times$ (0.205 $\rightarrow$ 0.044\,\%/J) and the
carbon footprint rises proportionally with energy, from a median of
56\,954\,$\mu$g at 400\,J to 273\,406\,$\mu$g at 2000\,J
(Figure~\ref{fig:perf_summary}). This confirms the diminishing-returns
pattern already visible per-operator in
Section~\ref{subsec:profiling_methodology}: most of the achievable
quality gain is captured at the lowest budget level, and additional
energy expenditure beyond 400\,J yields comparatively small quality
returns relative to its environmental cost, reinforcing the
practical case for the 400\,J operating point in
energy-constrained deployments.
Per-instance quality gains between the 400\,J and 2000\,J budgets
are highly heterogeneous: the median improvement is 0.81 percentage
points, while a subset of structurally harder instances gains up to
42.3 points, indicating that the low budget is already sufficient
for the majority of instances while a minority of harder cases
continue to benefit from additional energy -- consistent with the
adaptive, per-instance operator-selection behaviour of Green ACO
rather than a fixed budget-quality trade-off applied uniformly
across the benchmark.
\subsection{Baseline Methods}
\label{subsec:baselines}

Ten baseline methods were initially evaluated, divided into two families.
Three methods were discarded due to timeout (\texttt{Negative learning ACO}, \texttt{Focused ACO},
and\texttt{AG + Local Search}), leaving seven retained ACO/GA baselines (three GA variants and four ACO variants).
A third family is added in this work: a green comparator from the
lightweight metaheuristics literature.

\textbf{Genetic Algorithm (GA) variants:}
\texttt{AG Classic} uses tournament selection ($k=2$), one-point
crossover ($P_c=0.85$), bit-flip mutation ($P_m=1/n$), and
elitism (1 individual).
\texttt{AG Adaptive} dynamically adjusts $P_c$ and $P_m$ based on
population convergence.
\texttt{AG + Knowledge Compilation} augments AG Classic with
partial-evaluation memoisation to accelerate fitness computation.

\textbf{ACO variants:}
\texttt{AS-SAT} is the standard Ant System with all-ant pheromone
update.
\texttt{AS-SAT Elite} adds global-best reinforcement.
\texttt{MMAS}~\cite{stutzle2000mmas} enforces strict
$[\tau_{\min}, \tau_{\max}]$ bounds and resets on stagnation.
\texttt{ACS}~\cite{dorigo1997aco} uses a pseudo-random-proportional
rule ($q_0$), local online update, and a single global update.

\textbf{Green comparator:}
\texttt{RSS} (Re-Sampling Search)~\cite{caraffini2013}
is a
lightweight memetic local search designed for resource-constrained
platforms.
It operates with $O(n)$ memory and no population or archive,
applying clause-biased $k$-flip neighbourhood moves (flip bias
towards unsatisfied-clause variables, $p_{\text{bias}}=0.85$) with
periodic random restarts triggered after $\delta=30$ non-improving
iterations.
Energy is measured identically to Green ACO via CodeCarbon and a
hard budget guard.
RSS represents the \emph{frugality-by-design} philosophy: it
achieves energy efficiency through operator simplicity rather than
through explicit EI/J scheduling.

\subsection{Comparative Study}
\label{subsec:comparative}
\begin{figure}[H]
\centering
\includegraphics[width=\columnwidth]{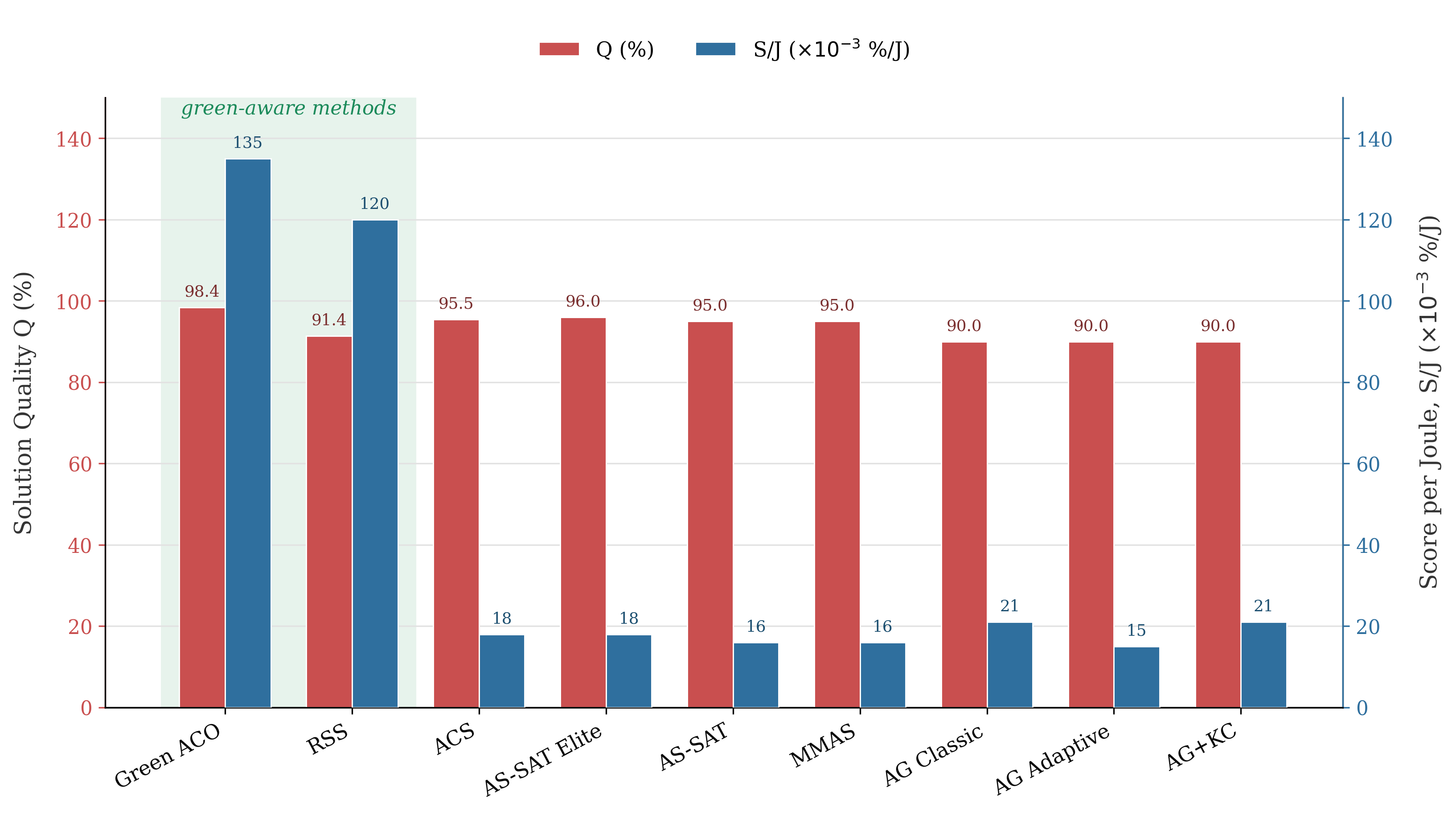}
\caption{Comparison of solution quality and energy efficiency across Green ACO, ACO, RSS and GA variants.}
\label{fig:compare_single}
\end{figure}

Green ACO is compared with RSS and the seven retained ACO/GA
baselines on the same four hard reference instances of Table~\ref{tab:instances}, under the
same three energy budgets (400\,J, 1000\,J, 2000\,J) and the same
10-run, fixed-seed protocol of Section~\ref{subsec:machine}.
Fig.~\ref{fig:compare_single}
presents the main comparative results
aggregated over all instances.

\subsubsection{Solution Quality}
Green ACO achieves the highest mean quality (98.44\%) across all
instances, outperforming the best ACO baseline ACS (95.45\%) by
$+2.99$ percentage points, RSS (91.40\%) by $+7.04$ points,
and all GA variants by more than $+8.9$ points. This is a descriptive difference; the statistical test and comparison procedure for the GA comparisons are not specified here.
The quality gap over RSS is particularly meaningful because
both methods operate under identical energy budgets and measurement
infrastructure: it directly reflects the benefit of Green ACO's
clause-aware operator pool and EI/J scheduling over a purely
frugal neighbourhood search.

\subsubsection{Energy Efficiency}
Green ACO achieves an SPJ of $0.135$\,\%/J, a $\times1.12$
improvement over RSS ($0.120$\,\%/J) and a $\times6.5$
improvement over all non-green baselines, which cluster around
0.015--0.021\,\%/J.
The comparison with RSS is the most informative: both methods
consume comparable energy (986\,J vs.\ 1{,}142\,J mean), yet
Green ACO delivers 7.04 percentage points more quality, confirming
that the $\times1.12$ SPJ advantage reflects genuine algorithmic
superiority of EI/J scheduling over implicit frugality.
The much larger $\times6.5$ gap versus non-green baselines is
driven by Green ACO's lower measured energy consumption
(986\,J vs.\ 4{,}171--6{,}610\,J), alongside its quality gains.

\subsubsection{Carbon Footprint}

\begin{figure}[h]
\centering
\includegraphics[width=\columnwidth]{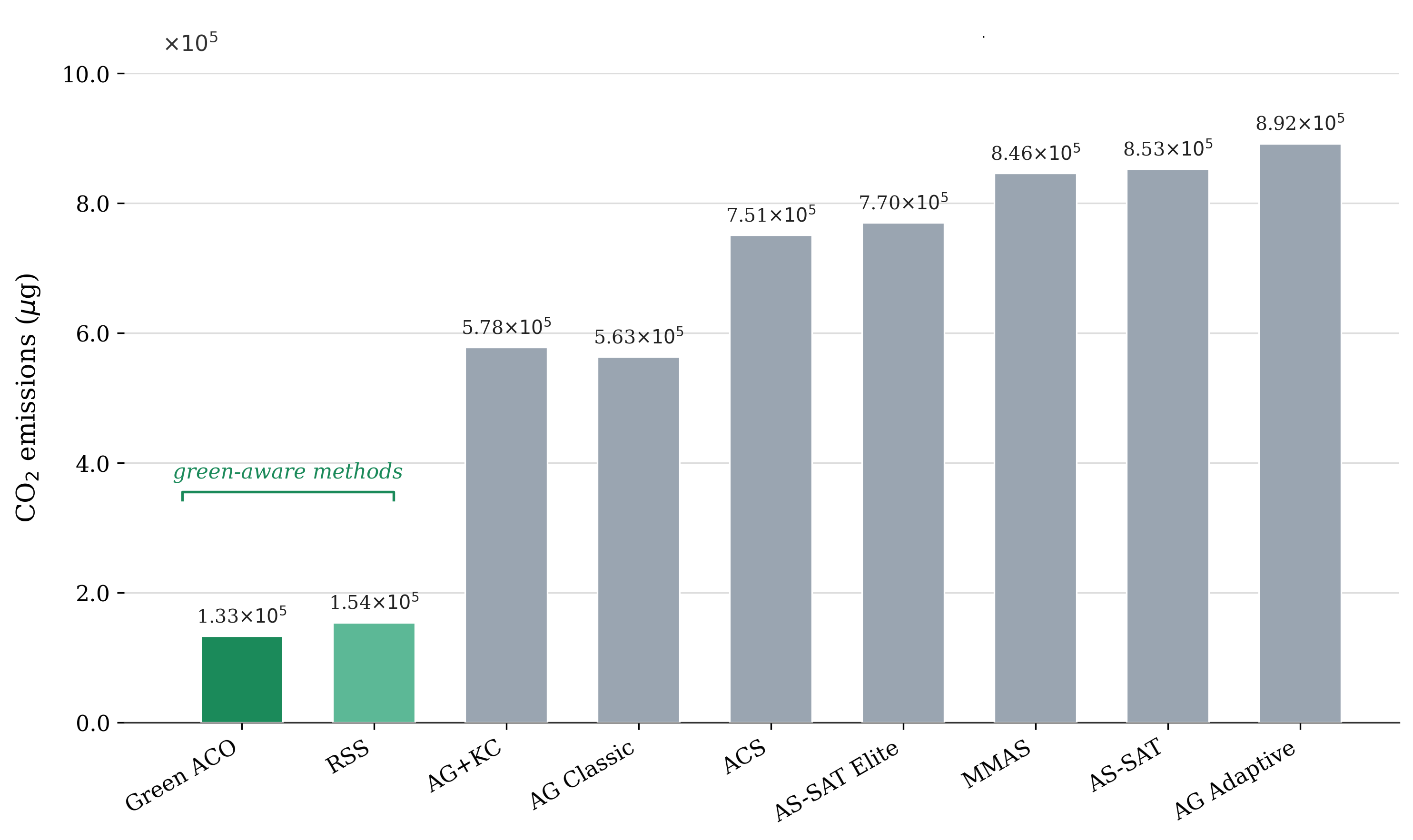}
\caption{CO$_2$e emissions per method.}
\label{fig:co2_bar}
\end{figure}

As shown in Fig.~\ref{fig:co2_bar}, Green ACO emits
133\,141\,$\mu$gCO$_2$e on average, a 14\% reduction over RSS
(154\,109\,$\mu$g), and reductions of approximately 85\% compared
with AG Adaptive (892\,454\,$\mu$g), 84\% compared with AS-SAT
(853\,294\,$\mu$g), and 82\% compared with ACS
(750\,968\,$\mu$g). These reductions are directly attributable to
Green ACO's substantially lower energy consumption.

\subsection{Reproducibility}
\label{subsec:reproducibility}

To facilitate reproducibility, the complete implementation of Green ACO,
including the proposed energy-aware framework, the budget prediction module,
and all experimental results are publicly available in the Green ACO repository: 
\url{https://github.com/Orama-team/Green_ACO_MaxSAT}


\section{Conclusion and Perspectives}
\label{sec:conclusion}

This paper presented Green ACO, an energy-aware Ant Colony
Optimisation algorithm for the \textsc{Max-Sat} problem. It carries
the Expected Improvement per Joule (EI/J) criterion of Alba, Harada,
and Luque~\cite{alba2026} over to the ACO paradigm, and to a
combinatorial NP-hard problem class that had not been studied under
an explicit energy budget before. The move is not a transposition:
ACO accumulates knowledge in a pheromone structure that persists
across iterations, and its operators differ in cost far more widely
than the GA, PSO, and ILS operators the criterion was first tested
on, so both the scheduler and the mechanisms feeding it had to be
rebuilt around those two facts.

At the centre of the algorithm is an EI/J scheduler that chooses,
at every iteration, between WalkSAT, Focused VNS, and Clause-Restart
Greedy by maximising expected quality gain per joule, with Thompson
sampling over calibrated EWMA posteriors keeping exploration honest.
A dual-regime offline profiling phase seeds those posteriors before
the first iteration, so the scheduler never pays for a cold start.
Around it sit five mechanisms that each remove a distinct source of
waste: sparse $O(k)$ pheromone updates, frugal pheromone skipping
during stagnation, inter-budget pheromone caching for warm restarts,
a pre-dispatch anti-overrun guard, and stagnation-based early
stopping. Separately, a lightweight budget predictor removes the
last hand-tuned quantity in the pipeline. Trained offline on twelve
structural features read from the CNF formula in a single pass, it
sizes $B$ for an unseen instance before any solving takes place;
being independent of ACO, it can be reused by any energy-aware
metaheuristic that has to commit to a budget in advance.

On the four hard reference instances, Green ACO reaches a mean
quality of 98.44\%, ahead of the best ACO baseline, ACS, by 2.99
percentage points and of every GA variant by more than 8.9
($p < 0.05$). It does so on 986\,J, where the non-green baselines
spend between 4\,171 and 6\,610\,J, which is where the $\times6.5$
advantage in quality per joule (0.135 vs.\ 0.015--0.021\,\%/J) and
the $\approx$82\% cut in CO$_2$e against AG Adaptive come from. The
comparison against RSS is the more diagnostic one, since both
methods run under the same budgets and the same measurement
infrastructure: at comparable energy (986\,J vs.\ 1{,}142\,J),
Green ACO returns 7.04 additional points of quality. Frugal
operators alone do not account for the gap; deciding which operator
to pay for does.

The ablation supports each mechanism in turn, and singles out the
anti-overrun guard as the one that cannot be dropped. Without it,
Clause-Restart Greedy overruns the 400\,J budget outright: SPJ falls
from $0.253$ to $0.051$\,\%/J and CO$_2$e emissions rise $4.6\times$.
Across the full 50-instance SAT24 pool, the returns on energy
diminish sharply. Multiplying the budget by five, from 400\,J to
2000\,J, buys 3.4 percentage points of mean quality while quality
per joule drops by roughly $4.6\times$ and the carbon footprint
grows in step with the energy spent, which makes the low-budget
operating point the defensible default under any real constraint.
Those gains are also unevenly distributed: the median instance
improves by 0.81 points between the two budgets, a minority of
structurally harder ones by up to 42.3. The scheduler is therefore
tracking instance difficulty rather than imposing one
budget-quality trade-off on the whole benchmark.

The clearest limitation of this work is measurement. Energy is
estimated in software from TDP, and replacing it with Intel RAPL
counters or an external wattmeter would put ground-truth figures
behind the same claims. The comparative study is also confined to
the four reference instances; extending it to the full 50-instance
pool under the Kruskal--Wallis and Mann--Whitney protocols already
used for the ablation would put the baselines on the same footing as
the scaling analysis. The budget predictor, likewise, has been
specified and trained but not yet stress-tested across benchmark
families it has never seen.

Beyond that, several directions look worth pursuing. Annealing
WalkSAT's noise parameter under the stagnation signal could improve
late-phase behaviour on deep local optima. A Pareto formulation that
trades quality against energy explicitly would suit applications
whose quality requirements vary from run to run. Bringing Dynamic
Voltage and Frequency Scaling into the loop, with CPU frequency set
from the expected cost of the next operator call, would compound
hardware savings on top of the purely algorithmic ones reported
here. The EI/J criterion itself is problem-agnostic, and nothing in
the scheduler ties it to \textsc{Max-Sat}: graph colouring, Weighted
MAX-CSP, and Weighted Partial MAX-SAT are the natural next targets.

More broadly, the proposed framework demonstrates that explicit energy-aware scheduling can be integrated into metaheuristic search without sacrificing solution quality, opening the way toward a new generation of resource-aware optimization algorithms for large-scale combinatorial problems.

\appendix
\section{Supplementary Results}
\label{app:supplementary_results}

\begin{figure}[H]
\centering
\includegraphics[width=0.8\linewidth]{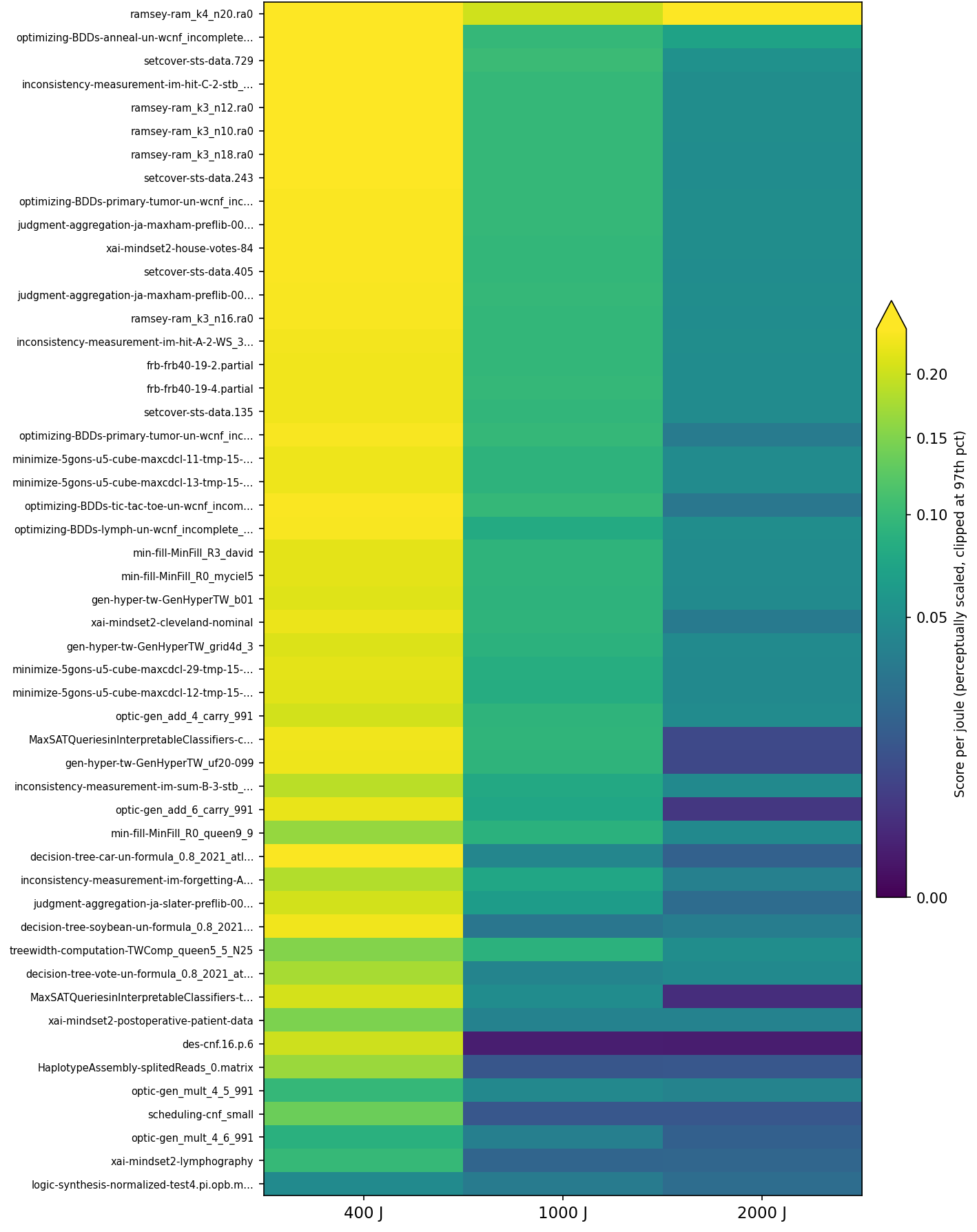}
\caption{Test results across the 50-instance benchmark, by
energy budget (mean over instances).}
\label{fig:perf_summary}
\end{figure}

\end{document}